\documentclass[journal]{IEEEtran}
\ifCLASSINFOpdf
\else
\fi
\usepackage{graphicx}
\usepackage[font=footnotesize,labelfont=bf,skip=0pt]{caption}
\usepackage{epstopdf}
\usepackage{amssymb}
\usepackage{amsmath} 
\usepackage{enumerate}
\usepackage{etaremune}
\usepackage{setspace}
\usepackage{comment}
\usepackage{color}
\usepackage[table]{xcolor}
\usepackage{soul,xcolor}
\setstcolor{red}
\usepackage{wrapfig}
\usepackage[algo2e]{algorithm2e} 
\usepackage{algorithm} 
\usepackage{algpseudocode} 
\usepackage[hidelinks]{hyperref}
\usepackage{cite}
\definecolor{blue}{RGB}{0, 111, 213}
\definecolor{blue0}{RGB}{100, 211, 100}
\definecolor{red}{RGB}{150, 11, 23}
\usepackage{soul}
\usepackage{booktabs}
\usepackage{float}
\usepackage{amsmath}
\usepackage{subfigure}
\usepackage{pgfplots} 
\pgfplotsset{compat=newest} 
\pgfplotsset{plot coordinates/math parser=false} 
\newlength\figureheight 
\newlength\figurewidth 
\usetikzlibrary{mindmap}
\usepackage{placeins}
\usetikzlibrary{arrows}
\usepgfplotslibrary{patchplots}

\definecolor{UBblue}{RGB}{0, 111, 213}
\definecolor{UBred}{RGB}{150, 11, 23}

 \usepackage{soul}
 \usepackage{epstopdf}

\def\subparagraph{}
\usepackage{threeparttable}

\begin{document}

\title{\LARGE{{\bf A Recurrent Neural Network Enhanced Unscented Kalman Filter for Human Motion Prediction}}}

\author{Wansong Liu$^{1}$, Sibo Tian$^{2}$, Boyi Hu$^3$, Xiao Liang$^{4}$, Minghui Zheng$^{2}$
    \thanks{This work was supported by the USA National Science Foundation  (Grants: 2026533, 2026276, and 2132923). This work involved human subjects or animals in its research. The authors confirm that all human/animal subject research procedures and protocols are exempt from review board approval.}
	\thanks{$^{1}$Wansong Liu is with the Mechanical and Aerospace Engineering Department, University at Buffalo, Buffalo, NY 14260, USA.
		{\tt\small Email: wansongl@buffalo.edu}.}%
  	\thanks{$^{2}$Sibo Tian and Minghui Zheng are with J. Mike Walker '66 Department of Mechanical Engineering, Texas A\&M University, College Station, TX 77840, USA.
		{\tt\small Emails: \{sibotian, mhzheng\}@tamu.edu}.}
	\thanks{$^{1}$Boyi Hu is with the Industrial and Systems Engineering Department, University of Florida, Gainesville, FL 32611, USA 
		{\tt\small Email: boyihu@ise.ufl.edu}.}%
	\thanks{$^{2}$Xiao Liang is with the Department of Civil \& Environmental Engineering, Texas A\&M University, College Station, TX 77840, USA.
    	{\tt\small Email: xliang@tamu.edu}.}
		\thanks{Correspondence to Minghui Zheng and Xiao Liang.}
}

\maketitle
\begin{abstract}
This paper presents a deep learning enhanced adaptive unscented Kalman filter (UKF) for predicting human arm motion in the context of manufacturing. Unlike previous network-based methods that solely rely on captured human motion data, which is represented as bone vectors in this paper, we incorporate a human arm dynamic model into the motion prediction algorithm and use the UKF to iteratively forecast human arm motions. Specifically, a Lagrangian-mechanics-based physical model is employed to correlate arm motions with associated muscle forces. Then a Recurrent Neural Network (RNN) is integrated into the framework to predict future muscle forces, which are transferred back to future arm motions based on the dynamic model. Given the absence of measurement data for future human motions that can be input into the UKF to update the state, we integrate another RNN to directly predict human future motions and treat the prediction as surrogate measurement data fed into the UKF. A noteworthy aspect of this study involves the quantification of uncertainties associated with both the data-driven and physical models in one unified framework. These quantified uncertainties are used to dynamically adapt the measurement and process noises of the UKF over time. This adaption, driven by the uncertainties of the RNN models, addresses inaccuracies stemming from the data-driven model and mitigates discrepancies between the assumed and true physical models, ultimately enhancing the accuracy and robustness of our predictions. One unique point of our method is that, it integrates a dynamic model of human arms and two RNN models, and uses Monte Carlo dropout sampling to quantify the uncertainties inherent in our RNN prediction models and transforms them into the covariances of the UKF's measurement and process noises respectively. Compared to the traditional RNN-based prediction, our method demonstrates improved accuracy and robustness in extensive experimental validations of various types of human motions. 

\end{abstract}


\section{Introduction}

The emergence of collaborative robots, which are typically designed to work side-by-side with human operators, has sparked a profound and revolutionary change in the domain of the production and manufacturing industry \cite{weiss2021cobots,lee2024review}. To foster a harmonious and safe human-robot partnership within the shared working space, robots should be equipped with the ability to understand and forecast their collaborator's intentions and behaviors so that robots can proactively adjust their motion to support human workers or avoid a potential collision. In this case, human motion prediction plays a crucial role in the next generation intelligent manufacturing system and several prior works have delved into this field within the context of human-robot collaborative assembly and disassembly \cite{liu2022dynamic, eltouny2023tgn, tian2023optimization, tian2023transfusion}.

Human motion prediction aims to anticipate potential movements of human agents within a given context or environment. It involves various techniques, spanning from traditional probabilistic model methods \cite{wang2005gaussian,ding2011human} to the latest trends rooted in deep learning. Deep learning techniques have been extensively applied to the human motion prediction problem in recent years, aiming at capturing the complex motion patterns exhibited by humans. Numerous studies employ Recurrent Neural Networks (RNNs) for sequence-to-sequence prediction due to the remarkable capacity of RNNs to capture temporal correlations in sequential data \cite{fragkiadaki2015recurrent,  martinez2017human, pavllo2018quaternet, jain2016structural}. Transformer \cite{aksan2021spatio, cai2020learning} and Graph Convolutional Networks (GCNs) \cite{dang2021msr, li2020dynamic} have also been applied in human motion prediction, offering the benefits of capturing the both spatial and temporal dependencies of human motion data.

\begin{figure*}[!]
	\centering 
	\includegraphics[scale=0.45]{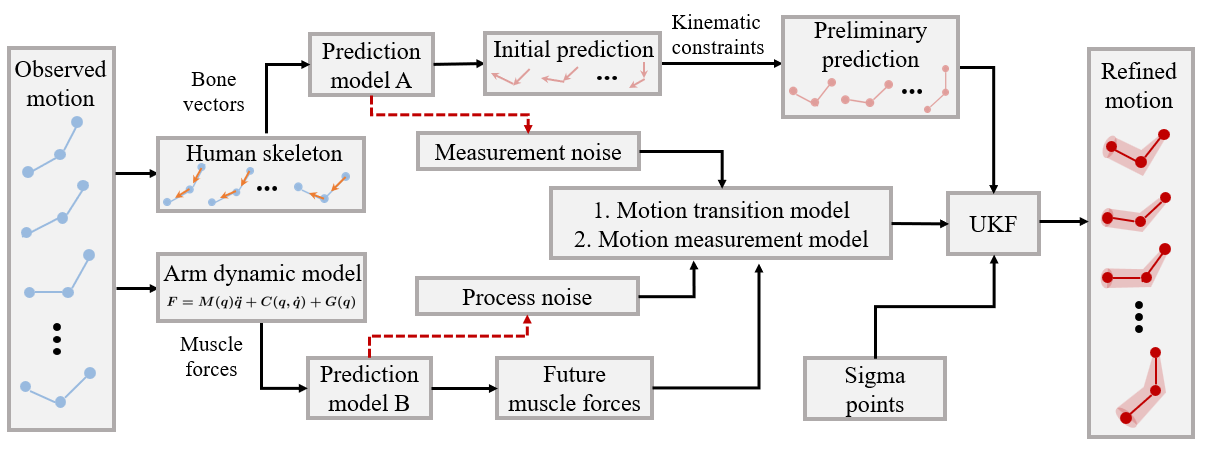}
	\caption{Overview of the proposed method: (1) We convert the observed motion into bone vectors and employ the prediction model A along with specific kinematic constraints like bone length to generate the preliminary prediction. The preliminary prediction serves as the measurement data of UKF. (2) Observed muscle forces are calculated based on the arm dynamic model, and the prediction model B is utilized to generate future muscle forces acting on the shoulder and elbow joints. (3) We quantify uncertainties of the prediction models A and B, and dynamically adjust the measurement and process noise covariances of UKF using the quantified uncertainties. (4) UKF eventually outputs the refined prediction, and the red shadow areas indicate uncertainties of refined motions.
	} \label{fig:overview}
\end{figure*}

Despite recent neural network-based models demonstrating good predictive capabilities in handling complex motion patterns, a notable issue remains. Current works usually involve a blind reliance on these black-box models, which excel at capturing intricate data relationships, and largely neglect the fundamental biomechanical principles, such as muscle forces and joint mechanics, that govern human motion. This disregard might lead to unrealistic or less precise predictions. It becomes evident that the prediction performance could be further enhanced by integrating neural networks with the underlying physics information of human agents. Several works have demonstrated the benefits of taking the physical information of the human body into consideration. For example, Lie algebra is employed to represent separate kinematic chains of the human body in \cite{liu2019towards, hu2019predicting, gui2018adversarial}, and the kinematic structure of the human skeleton is enhanced during the network training. Although Lie algebra-based approaches demonstrate impressive performance, the dynamics of human motion are still not incorporated in the neural networks, especially for the muscle force, which serves as a primary factor driving human motion.

Existing studies, such as \cite{li2019estimating,lv2016data,cao2015forecasting}, aim to establish a correlation between human motions and the corresponding muscle forces. For example, Lagrangian dynamic equations are used to characterize human body movements. These studies employ inverse dynamics to estimate the forces or torques of person-object interactions actuated by human. Meanwhile, a range of nonlinear estimation methods, including the particle filter \cite{chang20103d}, extended Kalman filter \cite{reif1999extended}, unscented Kalman filter (UKF) \cite{julier2004unscented}, and their modified versions, have been proposed to estimate the future state of human dynamic models. Among these, the UKF has demonstrated superior performance in balancing the accuracy and efficiency of state estimation \cite{gustafsson2011some}, and it has gained extensive usage in the realms of human motion tracking and prediction \cite{lee2016parallel,wang2017human,atrsaei2016human}. 

Traditional UKF-based methods assume constant and pre-defined measurement and process noises. However, such an assumption can result in estimation divergence as the characteristics of the noises evolve. Recently, adaptive filter algorithms have emerged, aiming to improve the accuracy and robustness of the state estimation. For example, in \cite{soltani2017improved}, different Kalman filters are fused by the ordered weighted averaging operator. This fusion is employed to iteratively update the noise covariance matrices. Rather than correcting the noise covariance during each iteration, the work presented in \cite{zheng2018robust} employs an online fault-detection mechanism. This mechanism decides the suitable times to generate new noise covariance matrices and replace the current ones.

In this paper, we consider a human arm model with three joints, including shoulder, elbow and wrist. We employ a physical model based on Lagrangian-mechanics to establish the intrinsic connection between human motions and muscle forces. Furthermore, we leverage an adaptive UKF to predict future human motions using this established connection. Fig.~\ref{fig:overview} illustrates the overview of our proposed method. Initially, we employ the data-driven prediction model A along with specific kinematic constraints, such as bone length, to derive a preliminary prediction. Different from traditional UKF-based prediction methods that require the real-time measurement data obtained by sensors, we consider this preliminary prediction as the measurement motion data used in the UKF, allowing the model to have a long prediction horizon. Subsequently, we compute observed muscle forces based on the arm dynamic model and utilize the prediction model B to estimate future muscle forces. By incorporating these future muscle forces, our motion transition model is capable of computing future arm motions. Eventually, the UKF is leveraged to obtain refined future motions.

Additionally, accurately defining the measurement and process noises of UKF poses a challenge. The measurement noise of UKF represents inaccuracies and errors of the measurement data, while the process noise of UKF implies variations of system dynamics that are not explicitly accounted by the model. Although the noise covariances can be manually tuned by users in \cite{liu2022dynamic}, the tuning process is resource-intensive due to the diverse and stochastic nature of human motions. In this study, we employ Monte Carlo dropout sampling (MCDS) method to explicitly quantify uncertainties of the prediction models A and B of Fig.~\ref{fig:overview}. Specifically, uncertainties of the data-driven prediction model A imply inaccuracies of the preliminary prediction. These uncertainties are naturally converted to the measurement noise covariance of UKF. Similarly, uncertainties of the physical model (i.e., the motion transition model of Fig.~\ref{fig:overview}), arising from the incorporation of future muscle forces, are transformed to the process noise covariance of UKF. In general, rather than relying on user-driven heuristic tuning for these two noise covariances, we integrate model uncertainties into the UKF framework. This integration allows an adaptive adjustment of the covariances during the prediction process. Such human motion prediction with possibility of explicitly quantifying uncertainties can be used in task sequence planning and robotic motion planning in human-robot interactive environments \cite{lee2022task,lee2022robot,liu2023task} to enable smoother and safer collaboration between human and robots.

\section{Prediction model and uncertainties}
In this section, we briefly introduce the human motion prediction problem definition and notations. Then we show how the network-based prediction model could be used to get the preliminary prediction as the measurement data of UKF, and present details of how we quantify uncertainties of the prediction model as the measurement covariance. 

\subsection{Prolem definition and RNN-based prediction model}
Human motion prediction seeks to forecast future movement sequence data based on the observed motion data, and is typically formed as a regression problem. In this study, we focus on analyzing arm's motion within the context of manufacturing. The human arm is represented by a three-joint skeleton, including the shoulder, elbow and wrist. We assume that the position of the shoulder is fixed, so the movement of the arm is represented by the positional changes of the elbow and wrist. Instead of using the cartesian coordinates of these two joints, we use unit bone vectors of upper arm and forearm to represent the arm pose, as it has the benefit of avoiding the relative translation issue of arm joints. Thus, an arm pose is denoted as $S=[s^a;s^b] \in \mathbb{R}^6$, where $s^a \in \mathbb{R}^3$ and $s^b \in \mathbb{R}^3$ respectively indicate unit bone vectors of the upper arm and forearm. The observed arm poses are denoted as $\mathbf{S}=[S_{-N+1},\dots,S_0] \in \mathbb{R}^{6 \times N}$, where $N$ is the observed step horizon. The future arm poses are denoted as $\hat{\mathbf{S}}=[\hat{S_{1}},\dots,\hat{S_M}] \in \mathbb{R}^{6 \times M}$, where $M$ is the predicted step horizon.

To obtain the prediction of human arm, we employ the long short-term memory (LSTM) as the prediction model. It excels at capturing long-term dependencies in sequential data, making them beneficial for understanding the context and relationships of the observed arm poses. The network takes the observed motion sequence $\mathbf{S}$ as inputs, and incorporates kinematic constraints, such as the fixed bone length, to re-construct arm poses during the training process. Thus the prediction process is denoted using the following equation:
\begin{equation}
    \hat{\mathbf{S}}=LSTM(\mathbf{S},\mathbf{\Theta})  \label{LSTM_S}
\end{equation}
where $\mathbf{\Theta}$ indicates the parameters of the LSTM model.

\begin{figure}[!]
	\centering 
	\includegraphics[scale=0.38]{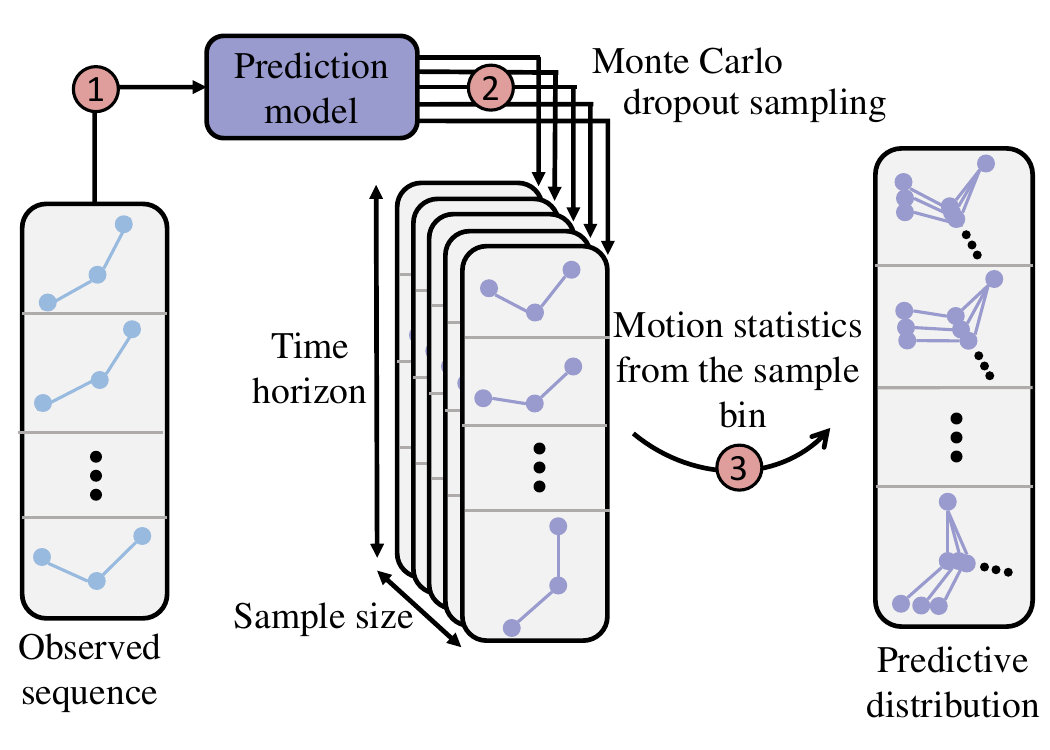}
	\caption{Uncertainty quantification: The observed arm poses are represented using blue color, the predicted arm poses are represented using purple color. Using Monte Carlo dropout sampling method, a single observed sequence can generate multiple predicted sequences. The predictive distribution is generated based on the motion statistics from the sample bin. Each step contains multiple possible arm poses.
	} \label{fig:uncertainty}
\end{figure}

\subsection{Uncertainty quantification of the prediction model}
Due to the inherent variability in human motions, blindly relying on the output of the prediction model is not advisable. Additionally, uncertainty is an intrinsic aspect of the prediction model. Therefore, rather than obtaining a deterministic prediction sequence from Eq.~(\ref{LSTM_S}), we explicitly measure the uncertainties of the prediction model and predict human motions probabilistically. 

The uncertainty exploited in this study arises from the parameters of the LSTM model, and provides insights into the confidence level associated with the model's output. Multiple methods attempt to quantify the uncertainty of the network-based model, such as variational autoencoders \cite{franchi2020encoding}, ensemble methods \cite{lakshminarayanan2017simple}, and MCDS methods \cite{gal2016dropout}. In this study, we choose MCDS to quantify uncertainties of prediction models considering it can accurately measure the uncertainty with less training efforts and computational costs. 
Dropout is conventionally employed to prevent overfitting in networks by randomly deactivating a set of network units.  
To quantify this uncertainty, we consider dropout in our LSTM model as the Bayesian approximation over the LSTM model parameters \cite{gal2016dropout}. The prediction distribution can be derived using the following equation:
\begin{equation}
p(\hat{\mathbf{S}}|\mathbf{S})=\frac{p(\hat{\mathbf{S}}|\mathbf{S},\mathbf{\Theta}) p(\mathbf{\Theta})}{p(\mathbf{\Theta}|\mathbf{S},\hat{\mathbf{S}})}
\end{equation}
where $p(\mathbf{\Theta})$ is a prior probability of the model parameters, $p(\hat{\mathbf{S}}|\mathbf{S},\mathbf{\Theta})$ stands for the likelihood used to capture the prediction process, and $p(\mathbf{\Theta}|\mathbf{S},\hat{\mathbf{S}})$ indicates the posterior probability distribution. Although the posterior distribution is intractable, we can approximate it with a distribution $q(\mathbf{\Theta})$ through variational inference \cite{graves2011practical}. This approximated distribution can be learned by minimizing the Kullback-Leibler divergence between $q(\mathbf{\Theta})$ and the actual posterior. 

Additionally, based on \cite{gal2015bayesian}, the training process of the prediction model is beneficial for learning $q(\mathbf{\Theta})$. Consequently, the predicted variance of arm motions $\mathbf{u}^{\hat{\mathbf{S}}}$ at test time using MCDS is denoted as \cite{kendall2017uncertainties}: 
\begin{equation}\label{uncertainty}
\mathbf{u}^{\hat{\mathbf{S}}} \approx \frac{1}{K} \sum_{k=1}^{K} LSTM(\mathbf{S},\overline{\mathbf{\Theta}}_k)^{T}
LSTM(\mathbf{S},\overline{\mathbf{\Theta}}_k)-E^{T} E
\end{equation}
where $\mathbf{u}^{\hat{\mathbf{S}}}=[u_1^{\hat{S}},...,u_M^{\hat{S}}]$ indicates uncertainties of the prediction model given the observed motion sequence $\mathbf{S}$, $K$ is the number of samples generated through random dropout during the evaluation stage, $\overline{\mathbf{\Theta}}_k$ stands for the model parameters of $k$th sample after dropout and is fitted to $q(\mathbf{\Theta})$, and $E \approx \frac{1}{K}{\textstyle\sum}_{k=1}^{K} LSTM(\mathbf{S},\overline{\mathbf{\Theta}}_k)$ indicates the predictive mean, which is used as the measurement data in UKF. Furthermore, the predictive variance $\mathbf{u}^{\hat{\mathbf{S}}}$ is converted into the measurement noise covariance in UKF. Fig.~\ref{fig:uncertainty} illustrates the uncertainty quantification of the prediction model. The observed motion sequence is first fed into the LSTM-based prediction model. Subsequently, the MCDS method is applied to generate multiple prediction samples during the evaluation phase. Finally, the predictive distribution is acquired based on the motion statistics from the sample bin. 

\section{Physics-informed human motion prediction}
In this section, we present an arm dynamic model based on the principles of Lagrangian-mechanics. This model serves to establish the relationship between arm motions and the corresponding muscle forces. Furthermore, we explain how to adaptively update the measurement and process noises of UKF and predict accurate arm motions. 

\begin{figure}[!]
	\centering 
	\includegraphics[scale=0.4]{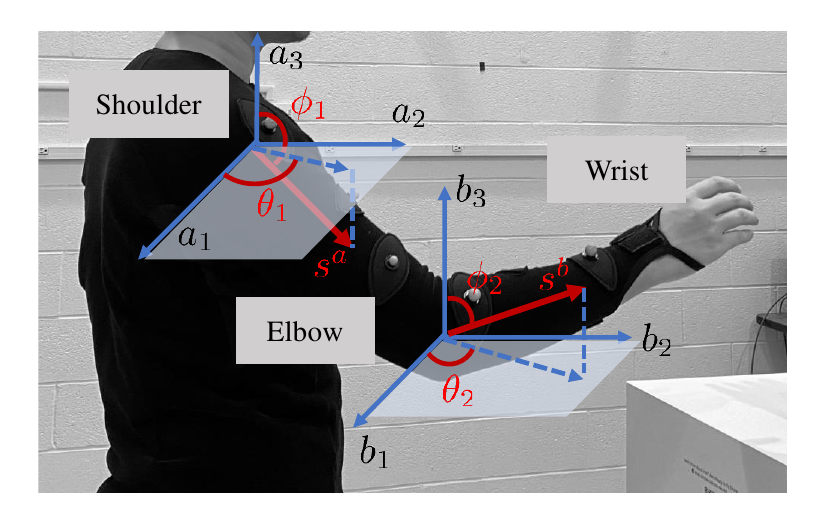}
	\caption{Human arm model: The arm motion is tracked using $\phi_1$, $\theta_1$, $\phi_2$, and $\theta_2$. Additionally, $\phi_1$ is the angle between $a_3$ and the vector of upper-arm $s^a$, $\theta_1$ is the angle between $a_1$ and the projection vector of $s^a$, $\phi_2$ is the angle between $b_3$ and the vector of forearm $s^b$, and $\theta_2$ is the angle between $b_1$ and the projection vector of $s^b$.
	} \label{fig:arm_model}
\end{figure}

\begin{figure}[!]
	\centering 
	\includegraphics[scale=0.27]{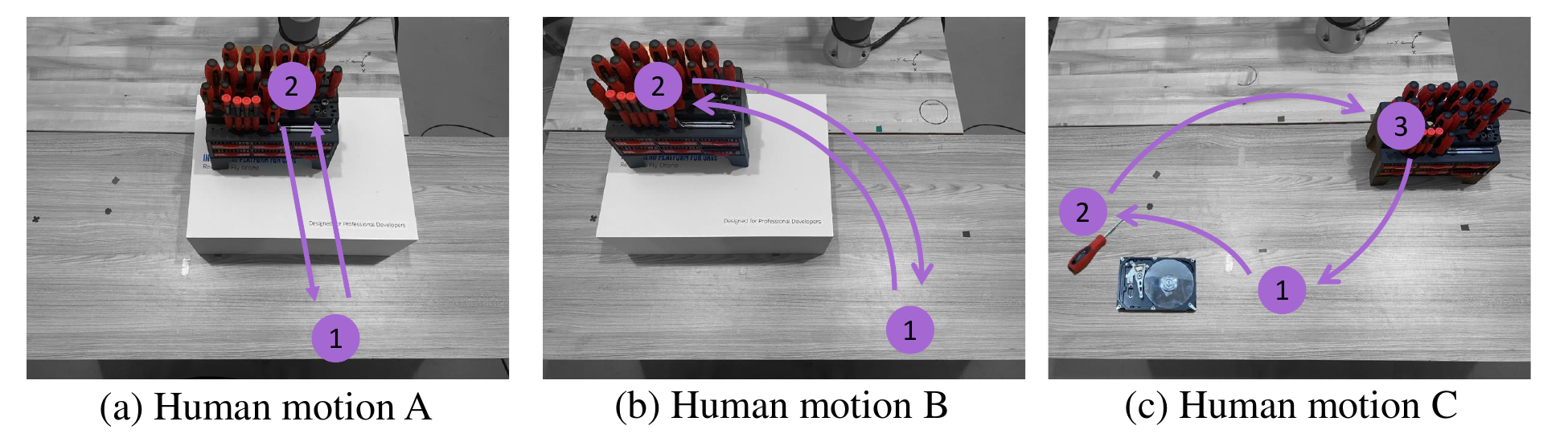}
	\caption{Illustration of three human motions: The numbers 1, 2, and 3 indicates the key positions of human hand. In motion A, the human worker first moves forward to grab a screwdriver in the toolbox, then moves back to the start position on the desk. In motion B, the human worker initially grabs a screwdriver on the left side, then moves back to the start position. In motion C, the human worker's first action is to pick up the screwdriver on the desk, followed by placing it into the toolbox, and ultimately returning to the start position. 
	} \label{fig:human_motions}
\end{figure}

\subsection{Arm dynamic model connecting motion and muscle force}
Fig.~\ref{fig:arm_model} illustrates the human arm model and two reference frames of the shoulder and elbow joints. We employ $\phi_1$ and $\theta_1$ to track the motion of the upper-arm, and $\phi_2$ and $\theta_2$ to track the motion of the forearm. We denote $q=[\phi_1;\theta_1;\phi_2;\theta_2] \in \mathbb{R}^{4}$ as the generalized coordinate, and consider the force $F$ acting on joints as the only generalized force in this study.
Euler-Lagrangian equations are used to describe the arm motion dynamics:
\begin{equation}
F=M(q) \ddot{q}+C(q, \dot{q})+G(q) \label{dynamicsmodel}
\end{equation}
where $M(q)$ is the inertia matrix, $C(q, \dot{q})$ is a velocity coupling matrix, $G(q)$ is a gravitational force vector. The observed muscle force sequence $\mathbf{F} \in \mathbb{R}^{4 \times N}$ are calculated based on Eq.~(\ref{dynamicsmodel}). Similarly, we employ the methods of motion prediction and uncertainty quantification in Section II to generate future muscle forces $\hat{\mathbf{F}}=[\hat{F}_{1},\dots,\hat{F}_M] \in \mathbb{R}^{4 \times M}$ and quantified uncertainties $\mathbf{u}^{\hat{\mathbf{F}}}=[u_1^{\hat{F}},...,u_M^{\hat{F}}]$. Note that the arm dynamic model also establishes the relationship between future arm motions $\hat{\mathbf{S}}$ and future muscle forces $\hat{\mathbf{F}}$.

\subsection{Adaptive unscented Kalman filter}
This section presents details of using an adaptive UKF to predict arm motions. The temporal state of human arm can be described based on the position and velocity of the arm joints. We use $x=[q;\dot{q}] \in \mathbb{R}^8$ to represent the state of the arm dynamic model. Eq.~(\ref{dynamicsmodel}) can be transformed to the discrete-time motion transition model:
\begin{equation}
\begin{aligned}
x_{m} &=x_{m-1}+G_{c}(x_{m-1},\hat{F}_{m-1}) T_{s} +\delta_m\\
&=G(x_{m-1},\hat{F}_{m-1})+\delta_m
\end{aligned}
\end{equation}
where $m$ indicates the index of the prediction step and the step horizon is $M$, the sampling time is denoted as $T_s$, and $\delta_m {\thicksim} (0,\rho u_{m}^{\hat{F}})$ is process noise in which $\rho$ is a scaling vector and $u_{m}^{\hat{F}}$ is the quantified uncertainty of the future muscle force. The motion measurement model is defined as:
\begin{equation}
y_{m}=H(x_{m})+\zeta_{m}
\end{equation}
where $H$ is a mapping function, and $\zeta_{m} {\thicksim} (0,\lambda u_{m}^{\hat{S}})$ is measurement noise in which $\lambda$ is a scaling vector and $u_{m}^{\hat{S}}$ is the quantified uncertainty of the future arm motion.

UKF employs the unscented transformation method \cite{julier1997new} to generate a set of sigma points $X$. The state mean $\hat{x}_m^{-}$ and covariance $P_m^x$ are calculated using the following equation:
\begin{equation}
\begin{aligned}
\hat{x}_m^{-} & = \sum_{i=0}^{2 L} W_{i} G(X_{m-1}^i,\hat{F}_{m-1}) \\
P_m^x & = \sum_{i=0}^{2 L} W_{i}\alpha \alpha^T +\rho u_{m}^{\hat{F}} \label{mc}
\end{aligned}
\end{equation}
where $L$ is the dimension of the arm state $x$, $W$ is assigned weights, and $\alpha=G(X_{m-1}^i,\hat{F}_{m-1})-\hat{x}_m^{-}$. In a similar vein, the measurement mean $\hat{y}_m^{-}$ and covariance $P_m^y$ are calculated using the following equation: 
	 \begin{equation}
    \begin{aligned}
    \hat{y}_m^{-} & = \sum_{i=0}^{2 L} W_{i} H(G(X_{m-1}^i,\hat{F}_{m-1})) \\
    P_m^y & = \sum_{i=0}^{2 L} W_{i}\beta \beta^T +\lambda u_{m}^{\hat{S}}. 
    \end{aligned}
    \end{equation}
where $\beta=H(G(X_{m-1}^i,\hat{F}_{m-1}))-\hat{y}_m^{-}$. The Kalman gain $K_m$ is calculated using the following equation:
\begin{equation}
    K_m=P_m^{xy} (P_m^y)^{-1}
\end{equation}
where $P_m^{xy}=\sum_{i=0}^{2L} W_i \alpha \beta^T$ indicates the cross co-relation matrix between the estimation and measurement. We convert the mean of the predicted motion sequence $E_m$ to $y_m^*$, which is treated as the sensor-based measurement data in UKF.  
Eventually, the state of arm dynamic model is predicted using the following equation:
\begin{equation}
\begin{aligned}
\hat{x}_m &= \hat{x}_{m}^-+K_m(y_m^*-\hat{y}_m^-)\\
\hat{P}_m^x &= P_m^x-K_m P_m^y K_m^{T}
\end{aligned}
\end{equation}
where $\hat{x}_m$ indicates the refined arm motion and $\hat{P}_m^x$ stands for the updated prediction covariance used for generating sigma points of the next iteration. The arm motion sequence $\hat{\mathbf{x}}=[\hat{x}_1,\dots,\hat{x}_m,\dots,\hat{x}_M]$ is recursively predicted.

\section{Experimental validations}
\subsection{Different motion datasets and network training}
To validate the effectiveness of our method, we develop human motions A, B, and C, as illustrated in Fig.~\ref{fig:human_motions}. In human motions A and B, the human worker grabs a screwdriver from the toolbox, located in the front of human worker and on the left side of human worker respectively. In human motion C, the human picks up a previously used screwdriver and returns it to the toolbox on the right side. We use the Vicon motion capture system to record 132 trajectories for motion A, 144 trajectories for motion B, and 152 trajectories for motion C in the frequency of 25 Hz. Note that one trajectory can generate multiple sets of observation and prediction sequences. 

By integrating the anthropometric data, such as bone length and inertial moments of upper arm and forearm, we compute joint muscle forces based on Eq.~(\ref{dynamicsmodel}). All trajectories are converted to unit bone vectors and muscle forces. 70\% of the converted data is allocated for training the prediction models, 15\% of them is used for validation, and the remaining portion is reserved for testing. The observation horizon $N$ and prediction horizon $M$ are both designed to be 50, which implies we use the preceding 2 seconds of data to forecast the subsequent 2 seconds of data. Addtionally, we use MCDS method with $K=10$ to quantify the uncertainties of the prediction models, and adjust the measurement and process noise covariances of UKF.

\begin{figure}[!htbp]
	\centering 
	\includegraphics[width=0.85\columnwidth]{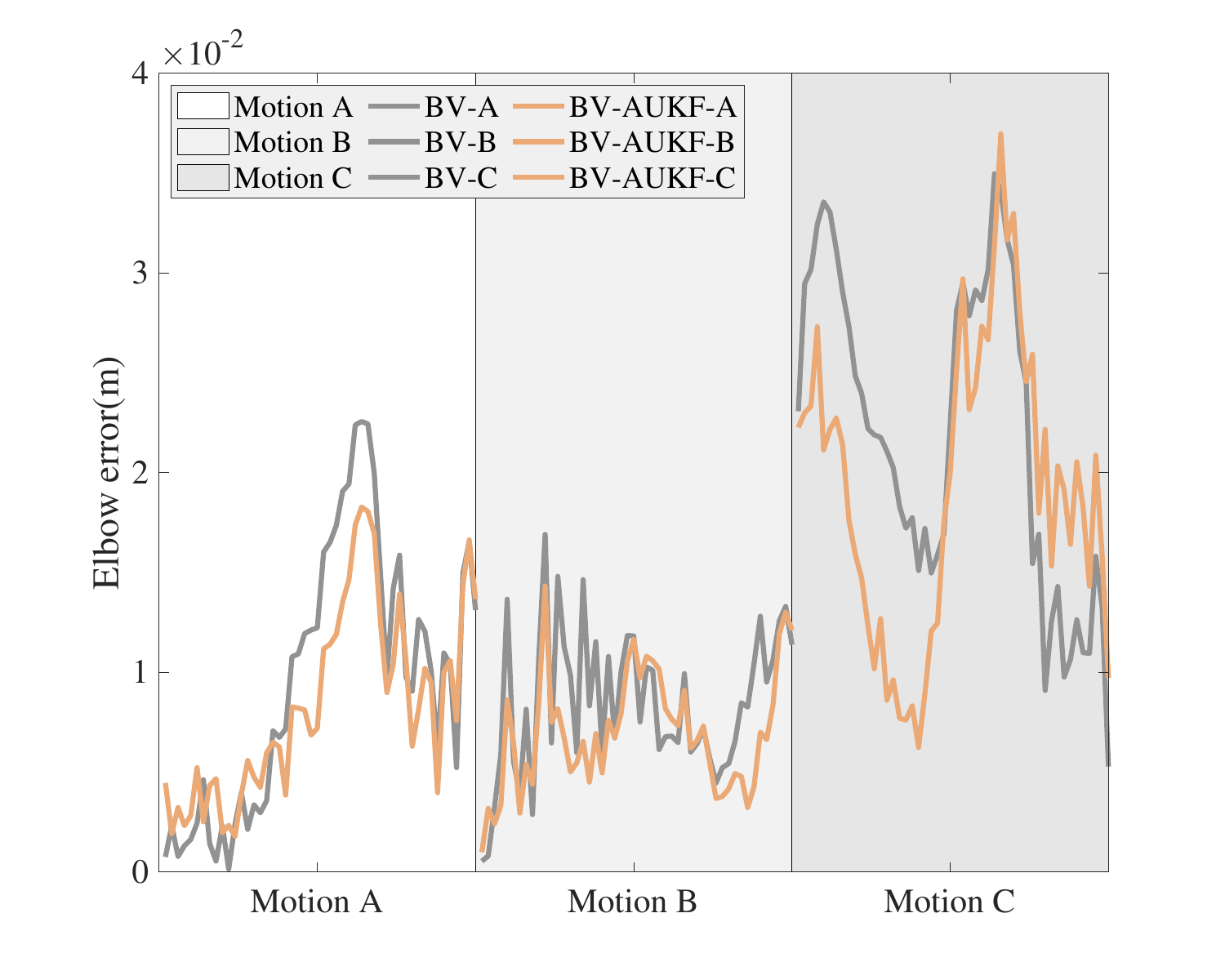}
	\caption{Prediction error of elbow using BV and BV-AUKF for each motion category. Each motion sample has 50 steps of prediction.  
	} \label{fig:elbow_error}
\end{figure}

\begin{figure}[!htbp]
	\centering 
	\includegraphics[width=0.85\columnwidth]{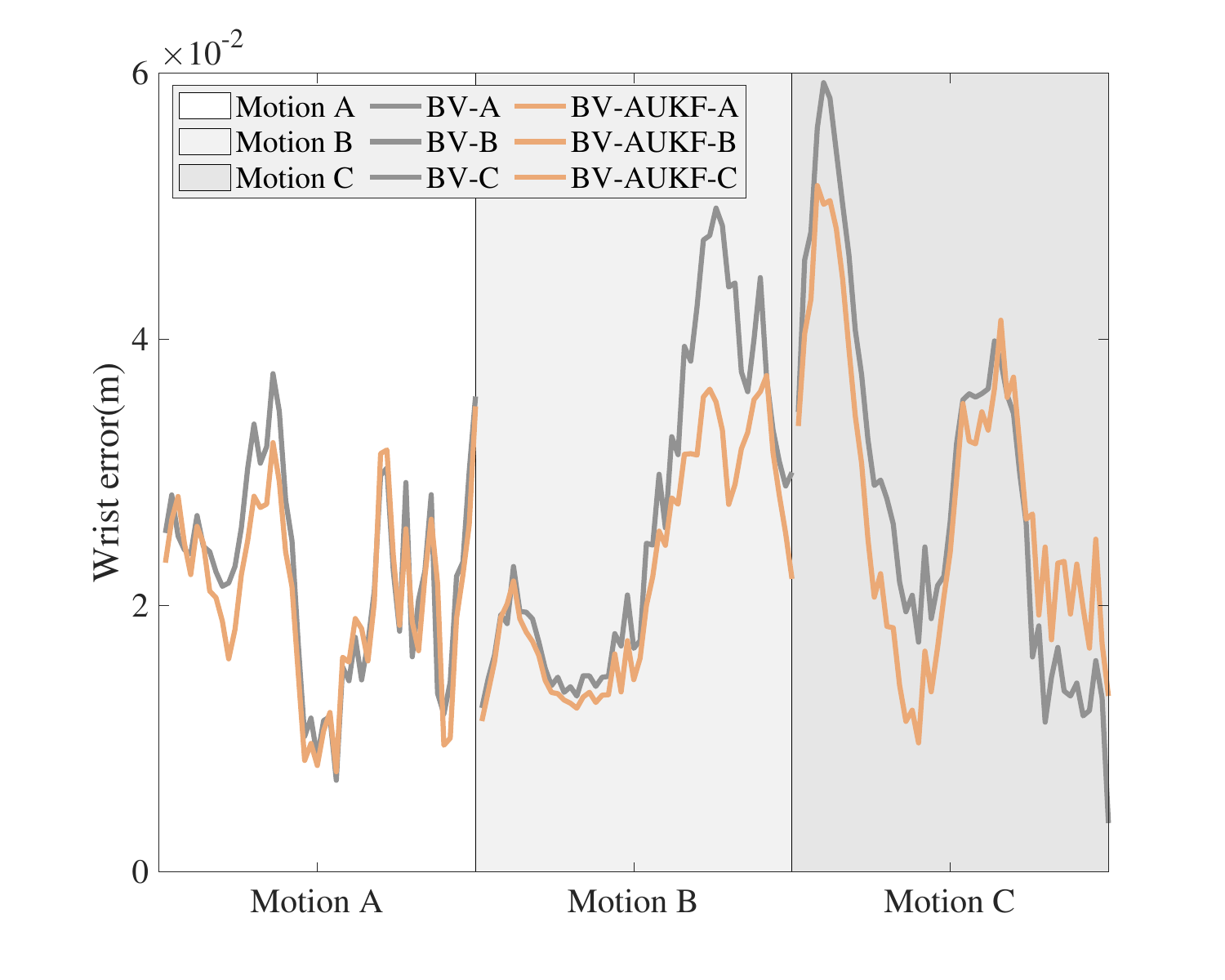}
	\caption{Prediction error of wrist using BV and BV-AUKF for each motion category. Each motion sample has 50 steps of prediction. 
	} \label{fig:wrist_error}
\end{figure}

\begin{figure}[!]
	\centering 	
	\subfigure[Prediction error difference of BV and BV-AUKF in motion A]{
	\includegraphics[width=0.9\columnwidth]{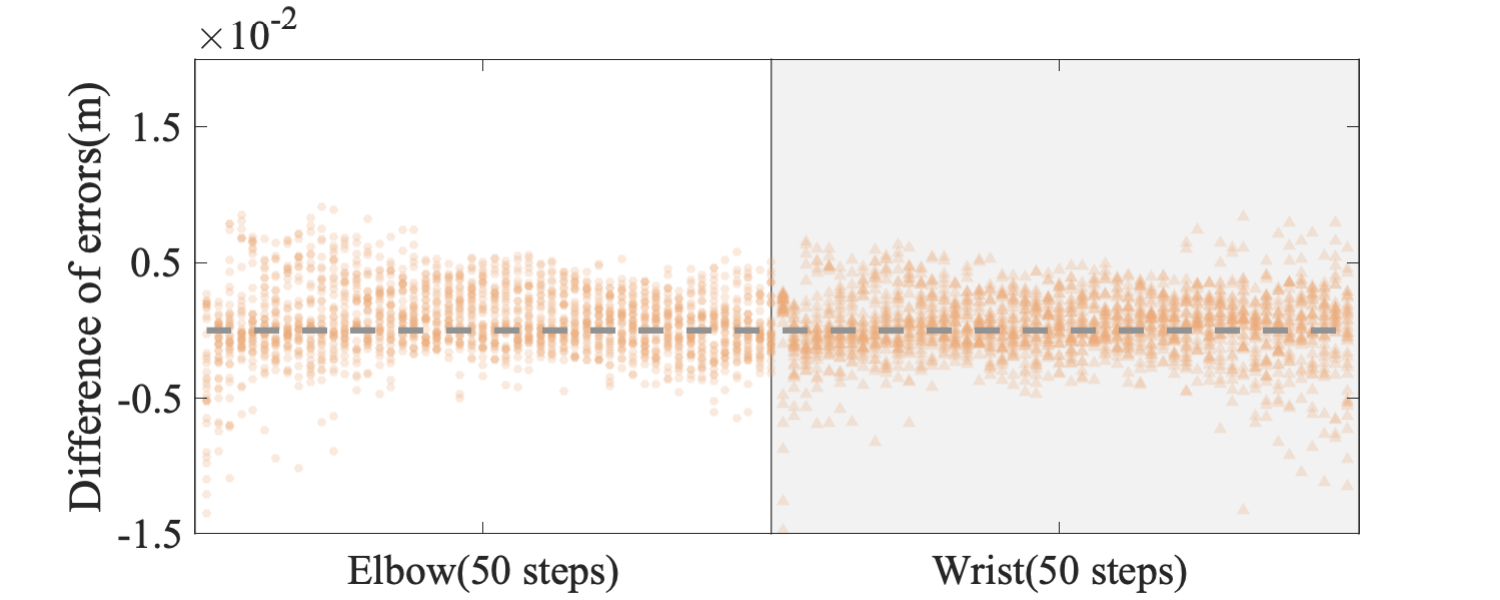}
	\label{fig:motion_a}
	}
	\subfigure[Prediction error difference of BV and BV-AUKF in motion B]{
	\includegraphics[width=0.9\columnwidth]{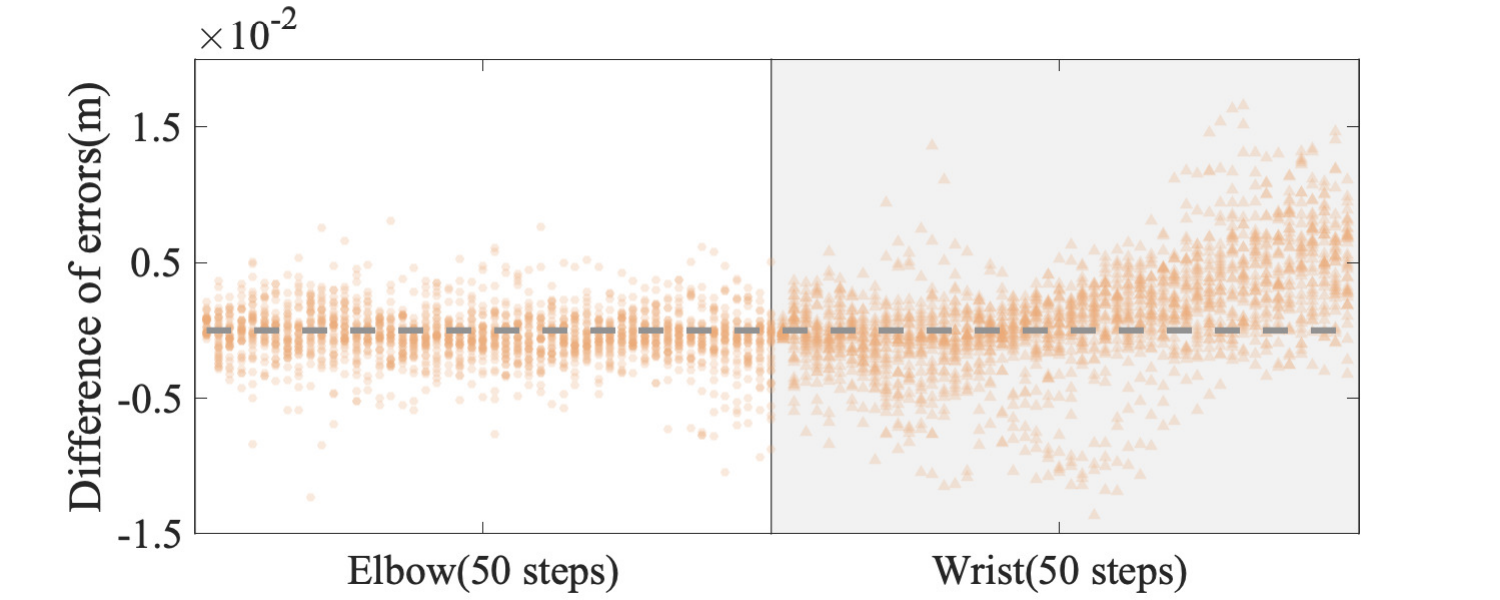}
	\label{fig:motion_b}
	}
 \subfigure[Prediction error difference of BV and BV-AUKF in motion C]{
	\includegraphics[width=0.9\columnwidth]{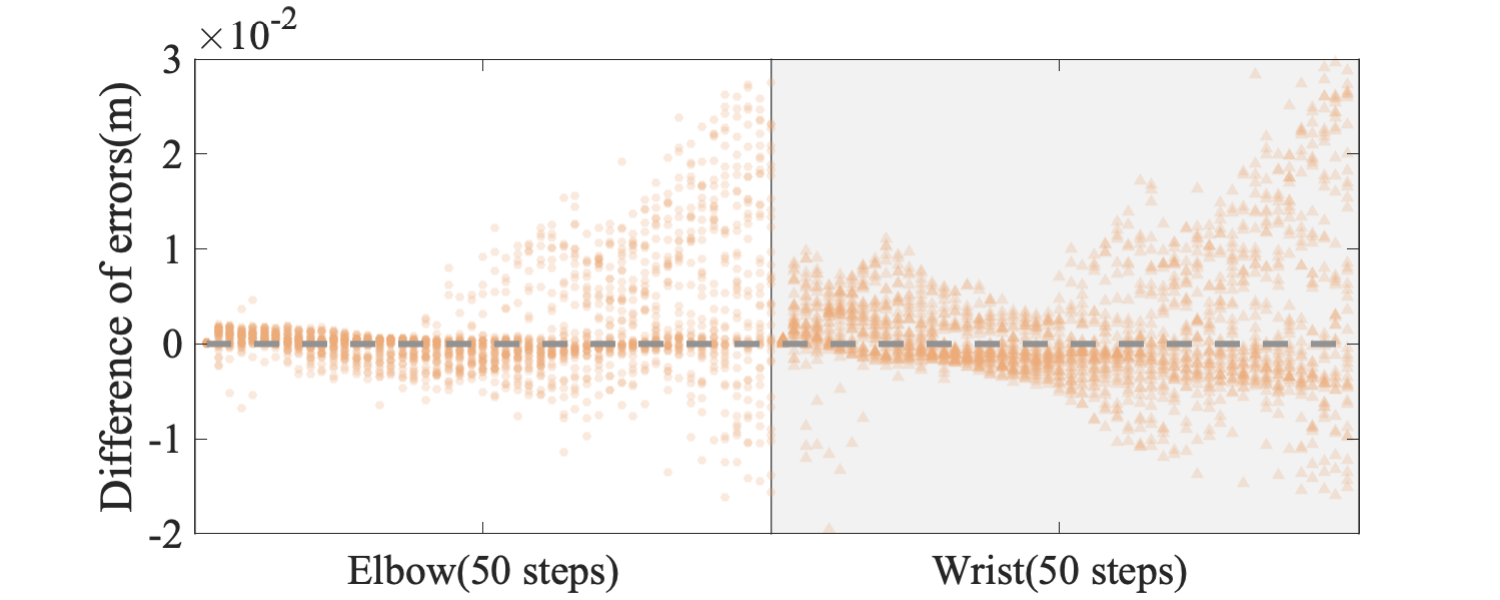}
	\label{fig:motion_c}
	}
	\caption{Prediction error difference using BV and BV-AUKF: We use the prediction errors based on BV to minus the prediction errors based on BV-AUKF. The gray dash line indicates zeros of the difference, which means equivalent prediction performance between two methods. The samples positioned above the gray dash line imply BV-AUKF outperforms BV in terms of prediction accuracy. The intensity of sample color indicates the number of overlapped samples. It shows the BV-AUKF performs better than BV in complex motions such as motion C and similar for simple motions such as motion A.}
	\label{fig:comparison}
\end{figure}

\subsection{Prediction results and discussion}
UKF iteratively provides the refined state of the arm dynamic model based on the whole state distribution. To better distinguish the prediction results using the traditional RNN and our method, we denote the traditional RNN-based prediction method using bone vectors as BV, and our physics-informed prediction method based on the adaptive UKF as BV-AUKF. Fig.~\ref{fig:elbow_error} and Fig.~\ref{fig:wrist_error} present the prediction errors of elbow and wrist at each time point, calculating from a single example motion in each motion category. Different blocks in the plots correspond to different human motions, with each motion sequence comprising 50 steps, equating to a 2-second duration of future arm motion. We use gray and orange colors to respectively present the prediction errors derived from BV and BV-AUKF. Based on the observation from the comparison, the orange lines are below the gray lines in the most of time, which implies that our method yields more accurate predictions in terms of the elbow and wrist positions.

\begin{table}[htbp]
\small
\caption{Quantitative results of error reduction percentage.}
\begin{center}
\begin{threeparttable}
\begin{tabular}{ccc|cc}
\toprule
            & \multicolumn{2}{c|}{Elbow}         & \multicolumn{2}{c}{Wrist}        \\ \cmidrule{2-5}
Motion      & AERP    & AMERP   & AERP   & AMERP \\ \midrule
A    & 5.28\%  & 43.90\% & 1.24\% & 22.00\% \\
B    & 0.13\%  & 36.48\% & 3.89\% & 34.65\& \\
C    & 3.82\%  & 34.95\% & 2.36\% & 24.98\% \\
\bottomrule
\end{tabular}
\begin{tablenotes}
     \item[*] AERP represents average error reduction percentage, AMERP represents average maximum error reduction percentage. The percentage is determined by dividing the difference in errors (error using BV minus error using BV-AUKF) by the prediction error using BV.
\end{tablenotes}
\end{threeparttable}
\end{center}
\label{tab:reduction}
\end{table}

To have a more comprehensive comparison, we randomly select 40 motion sequences for each type of motion from the test dataset. Each of these sequences has 50 steps. We then compute the prediction error differences by subtracting the prediction errors generated by BV-AUKF from those produced by BV, and present the error differences of three motions in Fig.~\ref{fig:comparison}. Note that the sample positioned above the gray dash line means that BV-AUKF has a smaller prediction error compared to BV. From Fig.~\ref{fig:comparison}, it is evident that the majority of samples are positioned above the red lines, indicating that the utilization of the adaptive UKF enhances the accuracy of the prediction for most arm poses. Another noteworthy observation is that within the motion C plot, a considerable number of samples reside above the he gray dash line, and many of them are significantly distant from it. Table.~\ref{tab:reduction} shows the advantage of our method in a more straightforward way. It presents quantitative results including the average error reduction percentage and the average maximum error reduction percentage of the prediction error from BV to BV-AUKF. Our method has a significant improvement with respect to the average of maximum error reduction of each sample. It is worth noting that, we have very carefully tuned the hyper-parameters of the RNNs in the BV human motion prediction method such that we can achieve the best possible results, while the covariance-related parameters in our BV-AUKF method are obtained in real-time with minimum tuning efforts. Therefore, even if the improvement is not significant, it is reasonable to state that the proposed RNN-enhanced adaptive UKF can improve the prediction of more dynamic or complex motions while having similar prediction performance for relatively simple motions compared to carefully tuned RNN models, and the tuning efforts of our method are minimal.

\section{Conclusions} 
This paper presents a recurrent neural network (RNN) enhanced unscented Kalman filter (UKF) to predict the motions of human arms. This method integrates a dynamic model of human arms and two RNN models. One RNN model provides a preliminary prediction that is treated as surrogate measurement data and fed into the UKF as the future measurement; the other RNN model predicts the muscle force of the human arms and is fed into the UKF as the future input. One unique point of our method is that, it uses Monte Carlo dropout sampling to quantify the uncertainties inherent in our RNN prediction models and transform them into the covariances of the UKF's measurement and process noises respectively. UKF adjusts measurement and process noises and their covariances in real-time. Experimental studies show that the proposed RNN-enhanced adaptive UKF, compared to very carefully tuned RNN models, can improve the prediction of more dynamic motions while having similar performance for relatively simple motions. Moreover, the real-time adaption of the covariances in UKF alleviates the tuning efforts of users.

\ifCLASSOPTIONcaptionsoff
  \newpage
\fi

\bibliographystyle{IEEEtran}
\bibliography{ref}{}

\end{document}